\documentclass[10pt,twocolumn,letterpaper]{article}

\usepackage[pagenumbers]{wacv} 

\definecolor{wacvblue}{rgb}{0.21,0.49,0.74}
\usepackage[pagebackref,breaklinks,colorlinks,allcolors=wacvblue]{hyperref}
\usepackage{multirow}
\def\wacvPaperID{1408} 
\def\confName{WACV}
\def\confYear{2026}

\title{Histopath-C: Towards Realistic Domain Shifts for Histopathology Vision-Language Adaptation}

\author{%
    \begin{tabular}[t]{c}
        Mehrdad Noori\thanks{Equal contribution. \\Correspondence to \href{mailto:mehrdad.noori.1@ens.etsmtl.ca}{mehrdad.noori.1@ens.etsmtl.ca}} \and
        Gustavo A. Vargas Hakim$^*$ \and
        David Osowiechi$^*$ \and 
    \end{tabular} \\
    \begin{tabular}[t]{c}
        Fereshteh Shakeri \and
        Ali Bahri \and
        Moslem Yazdanpanah \and
        Sahar Dastani
    \end{tabular} \\
    \begin{tabular}[t]{c}
         Ismail Ben Ayed \and Christian Desrosiers
    \end{tabular} \\[1em]
    \multicolumn{1}{p{\linewidth}}{\centering LIVIA, \'ETS Montreal, Canada \\ International Laboratory on Learning Systems (ILLS)}
}

\usepackage{amsmath,amsfonts,bm}
\usepackage[dvipsnames]{xcolor}

\newsavebox\CBox
\def\textBF#1{\sbox\CBox{#1}\resizebox{\wd\CBox}{\ht\CBox}{\textbf{#1}}}

\def\eqref#1{equation~\ref{#1}}

\def\1{\mathbf{1}}

\def\vt{{\mathbf{t}}}

\def\vx{{\mathbf{x}}}

\def\vz{{\mathbf{z}}}

\def\mA{{\mathbf{A}}}
\def\mB{{\mathbf{B}}}

\def\mP{{\mathbf{P}}}

\def\mS{{\mathbf{S}}}

\def\mW{{\mathbf{W}}}

\def\mY{{\mathbf{Y}}}
\def\mZ{{\mathbf{Z}}}

\DeclareMathAlphabet{\mathsfit}{\encodingdefault}{\sfdefault}{m}{sl}
\SetMathAlphabet{\mathsfit}{bold}{\encodingdefault}{\sfdefault}{bx}{n}

\DeclareMathOperator*{\argmax}{arg\,max}

\RequirePackage{xspace}
\makeatletter
\DeclareRobustCommand\onedot{\futurelet\@let@token\@onedot}
\def\@onedot{\ifx\@let@token.\else.\null\fi\xspace}

\def\eg{\emph{e.g}\onedot} 

\def\ie{\emph{i.e}\onedot}

\makeatother

\usepackage{booktabs} 
\usepackage{subcaption}
\usepackage{multirow}
\usepackage[table]{xcolor}

\newcommand{\mypar}[1]{\vspace{4pt}\noindent\textbf{#1~}}

\newcommand{\ppm}{\,\scriptsize$\pm$}

\newcommand{\phz}{\phantom{0}}

\newcommand{\improv}[1]{%
  \textcolor{OliveGreen}{\scriptsize{\, (+#1)}}%
}

\definecolor{SoftRed}{RGB}{200,60,60}
\newcommand{\degrad}[1]{%
  \textcolor{SoftRed}{\scriptsize{\, (-#1)}}%
}

\begin{document}
\maketitle
\begin{abstract}
Medical Vision-language models (VLMs) have shown remarkable performances in various medical imaging domains such as histo\-pathology by leveraging pre-trained, contrastive models that exploit visual and textual information. However, histopathology images may exhibit severe domain shifts, such as staining, contamination, blurring, and noise, which may severely degrade the VLM's downstream performance. In this work, we introduce Histopath-C, a new benchmark with realistic synthetic corruptions designed to mimic real-world distribution shifts observed in digital histopathology. Our framework dynamically applies corruptions to any available dataset and evaluates Test-Time Adaptation (TTA) mechanisms on the fly. We then propose LATTE, a transductive, low-rank adaptation strategy that exploits multiple text templates, mitigating the sensitivity of histopathology VLMs to diverse text inputs. 
Our approach outperforms state-of-the-art TTA methods originally designed for natural images across a breadth of histopathology datasets, demonstrating the effectiveness of our proposed design for robust adaptation in histopathology images.
Code and data are available at \href{https://github.com/Mehrdad-Noori/Histopath-C}{https://github.com/Mehrdad-Noori/Histopath-C}.
\end{abstract}
    
\section{Introduction}
\label{sec:intro}

\begin{figure}[th!]
    \centering
    \includegraphics[width=\linewidth]{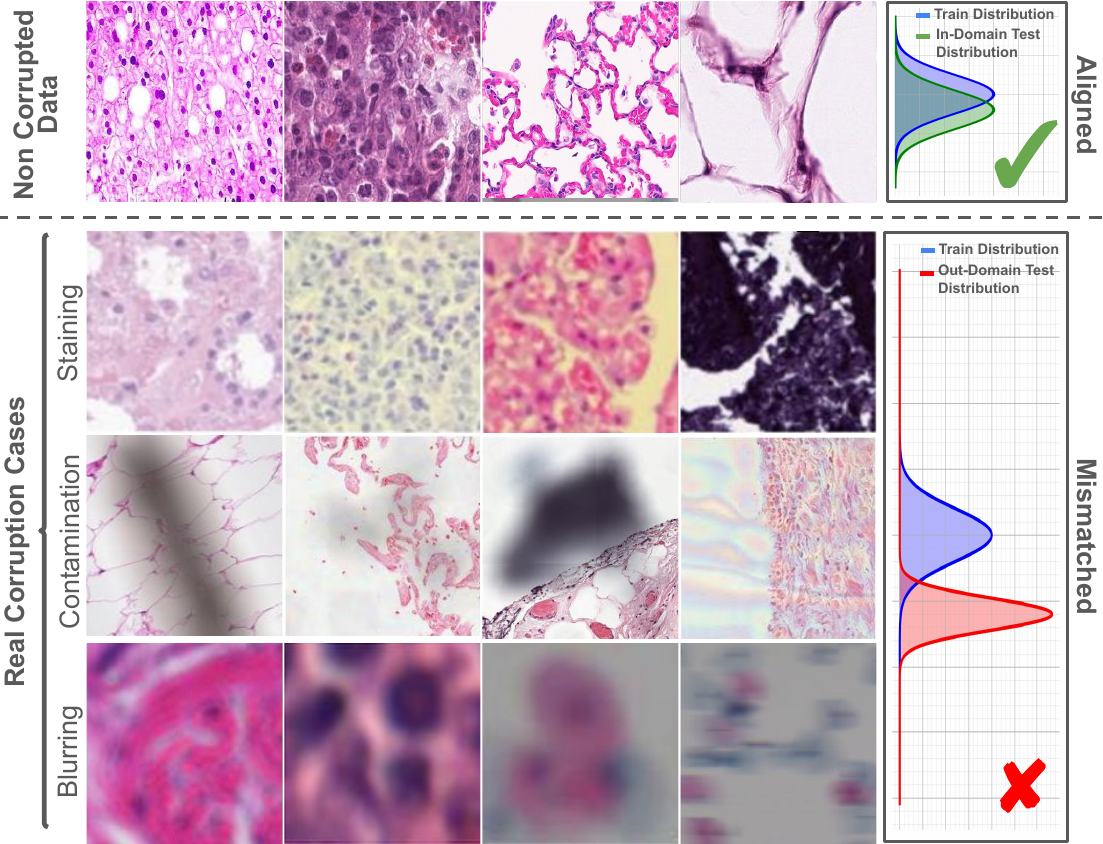} 
    \caption{Illustrative examples of real-world corruption artifacts in histopathology slides, as documented in prior literature. The top row shows representative clean (non-corrupted) images with train–test alignment. The bottom rows depict real instances of staining~\cite{stain_images, stain_2, stain_3}, contamination~\cite{contamination_image_1, contamination_image_2}, and blurring artifacts~\cite{blurring_image_4,blurring_image_5,blurring_image_6}, each introducing domain shifts that can significantly degrade model performance. These examples motivate the need for a test-time adaptation benchmark, such as Histopath-C, that simulates realistic perturbations and enables evaluation of adaptation methods in histopathology.}

    \label{fig:motivationm}
\end{figure}


Vision-Language Models (VLMs)~\cite{clip,align} have recently emerged as a powerful paradigm for solving a wide range of downstream visual tasks using a single large-scale pretraining objective that aligns visual and textual representations in a shared embedding space. While originally developed for natural image domains, VLMs have shown remarkable versatility and have been rapidly extended to medical imaging applications\cite{quilt,medclip,flair}. By leveraging large-scale paired image–text datasets—often sourced from clinical reports or curated descriptions—these models are able to learn transferable representations that outperform traditional convolutional or transformer-based architectures trained from scratch or with supervised objectives. Their strong zero-shot and few-shot generalization capabilities make them particularly appealing for medical domains, where labeled data is scarce and task diversity is high.


Despite their promising capabilities, medical VLMs inherit a critical limitation observed in standard computer vision models trained on datasets such as ImageNet~\cite{imagenet}: a pronounced sensitivity to domain shifts. In real-world clinical settings, variations in acquisition protocols, scanners, staining procedures, or patient populations are very common~\cite{contamination_image_2} and can lead to substantial distributional discrepancies between training and deployment data, which degrade the performance and reliability of deep learning models in clinical practice~\cite{stain_2, aubreville2021quantifying, blurring_image_3}. To mitigate domain shift, Test-Time Adaptation (TTA)~\cite{tent} has emerged as a compelling paradigm, enabling models to adapt at inference time by leveraging unlabeled target samples. While recent work has begun to explore TTA for VLMs~\cite{tpt,clipartt}, these efforts have been largely confined to natural image domains and have yet to be formally extended or systematically evaluated in the context of medical imaging.

To address this gap, we introduce a novel benchmark that simulates realistic domain shifts commonly encountered in histopathology. These shifts include staining, contamination, blurring, noise, illumination variations,. All corruptions can be applied on the fly to existing datasets, enabling systematic evaluation of model robustness under controlled distributional perturbations. Representative examples are shown in Figure~\ref{fig:motivationm}. While recent benchmarks such as Histo-VL~\cite{Histovl} aggregate heterogeneous datasets to study broad generalization and prompt/adversarial sensitivity, our focus is factor-isolated robustness. Histopath-C applies controlled, graded corruptions (stain, contamination, blur, illumination, noise) to the same images, allowing us to pinpoint which real-world artifacts most degrade each model and to stress-test TTA methods head-to-head under identical conditions.

We show that multiple pathology VLMs such as Quilt~\cite{quilt}, PathGen~\cite{pathgen}, and CONCH~\cite{conch} degrade substantially under these corruptions, and that existing TTA methods exhibit inconsistent performance across datasets and corruption types. This brittleness stems in part from the reliance of medical VLMs, on text templates derived from clinical reports, which can introduce high variability depending on the choice of prompt and input image.
To overcome these limitations, we propose a novel TTA framework that leverages multiple text templates to stabilize model predictions and enhance robustness under distribution shift. By aggregating information across diverse prompts, our method improves consistency and reduces sensitivity to template selection, a key weakness in existing VLM-based pipelines.
Our main contributions are summarized as follows:
\begin{itemize}
    \item A novel and challenging benchmark comprising 10 diverse domain shifts that closely resemble real-world corruptions encountered in histopathology imaging. We devise a framework for applying these corruptions \emph{on the fly} to any existing dataset.
    \item An evaluation of recent TTA methods for VLMs, demonstrating that the proposed benchmark introduces significant challenges that severely degrade performance, which existing baselines fail to address.
    \item We introduce Low-rank Adaptation with Transductive Template Ensembling (LATTE), a simple yet effective adaptation method for VLMs. This technique addresses text template ambiguity in histopathology imaging by leveraging loss-level aggregation across templates, transductive pseudolabeling, and low-rank adaptation.
    \item In addition to improving robustness under domain shift, LATTE consistently enhances zero-shot performance even in the absence of corruptions—demonstrating its effectiveness as a general-purpose enhancement for VLMs in medical imaging.
\end{itemize}

\begin{figure*}[t!]
    \centering
    \includegraphics[width=\linewidth]{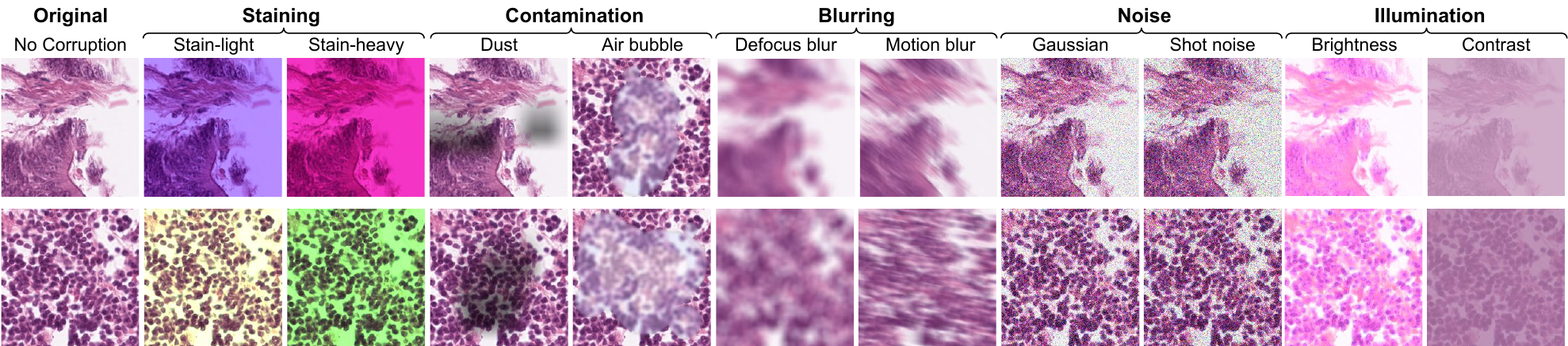}
    \caption{Representative examples of the ten corruption types introduced in the Histopath-C benchmark, spanning five categories: Staining, Contamination, Blurring, Noise, and Illumination. These synthetic corruptions are designed to mimic real-world perturbations in histopathology and can be dynamically applied to any dataset for robust evaluation.}
    \label{fig:benchmark}
\end{figure*}

\section{Related Work}
\label{sec:related}

\mypar{Medical Vision Language Models.} Building on the success of CLIP~\cite{clip}, significant efforts have been made to develop vision-language models (VLMs) for medical applications. These models have proven valuable in diverse fields, including retinal analysis \cite{flair}, radiology \cite{medclip}, and histopathology \cite{quilt}, by leveraging both visual and textual modalities. Their success largely stems from the availability of extensive visual data paired with detailed text descriptions \cite{pubmed}, along with powerful contrastive learning techniques that enhance the robustness of visual encoders. Despite their strong performance in downstream tasks, research has focused on adapting VLMs to new challenges with minimal supervision, such as few-shot learning \cite{fewshotvlms,promptsmooth}. Given the inherent scarcity of medical images, we argue that a zero-shot adaptation approach is more suitable, aligning with TTA, where the model continuously learns from each new sample in an unsupervised manner.


\mypar{Test-Time Adaptation.} In traditional TTA, a deep model is adapted \emph{on the fly} to new target domains that differ from the source training domain. Key aspects of this setting include the absence of labels during adaptation, the inaccessibility of the source dataset, and adaptation to data streams (\ie, batches) without access to the entire target dataset. In the spirit of exploiting the connection between the model's pre-training loss (\emph{e.g.}, crossentropy) and an unsupervised adaptation loss, TENT~\cite{tent} introduces conditional entropy minimization, and targets normalization layers' affine parameters. An important body of research has built on this methodology, offering increasing performance scores \cite{eta,conjugate,sar,sotta,stamp}. Alternative directions have been also investigated, including parameter-free Laplacian optimization~\cite{lame}, contrastive learning~\cite{cotta,adacontrast}, and pseudolabeling~\cite{shot,tast}. Test-Time Training (TTT) methods, such as \cite{ttt,tttflow,ncttt,clust3,recttt}, train an unsupervised sub-branch during pretraining for refinement during testing. 

TTA for VLMs has broader implications, as it enables model adaptation to any new downstream task, regardless of its domain. TPT~\cite{tpt} extended the framework of TENT by using entropy minimization on confident image augmentations' predictions, and using text prompt finetuning. However, this approach has inherent limitations in VLMs, as entropy minimization can lead to model degradation by adapting based on wrong and overconfident predictions. Additionally, the interactions between visual and textual modalities is not directly exploited. In response, recent methods have explored image-text pseudolabeling \cite{clipartt}, weight ensembling \cite{watt}, and intermediate features \cite{mlmp} in hopes of leveraging both modalities. However, as shown in Section~\ref{sec:experiments}, the improvements from these methods remain limited and inconsistent across the various domain shifts in our benchmark. Specifically, their working mechanisms appear to be suboptimal in the histopathology adaptation scenario.

\section{Benchmark}
\label{sec:benchmark}
We propose Histopath-C, a new TTA benchmark for histopathology imaging. Our setting is based on the application of ten different corruptions that realistically simulate real-world perturbations. Different from natural images~\cite{cifar10c}, our perturbations are specifically designed to reflect failure modes relevant to histopathology imaging. Moreover, Histopath-C can be applied to any dataset to evaluate the robustness of foundation models and adaptation algorithms. We categorize the chosen corruptions as five groups —\emph{Staining}, \emph{Contamination} (dust \& air bubbles), \emph{Blurring}, \emph{Noise}, and \emph{Illumination}. Representative examples of several important real-world artifacts, drawn from prior pathology studies, are shown in Figure~\ref{fig:motivationm}. We next outline how each corruption is simulated to reflect real-world variation.

\subsection{Staining}
Variations in staining procedures are one of the most pervasive and impactful sources of domain shift in histopathology imaging \cite{stain_images, stain_2, stain_3}. These variations can arise from differences in reagent concentration, staining protocols, scanner calibration, or tissue fixation practices across laboratories and institutions. The resulting shifts in color composition and intensity patterns (as shown in Figure~\ref{fig:motivationm} can drastically alter the appearance of tissue structures and reduce model generalization, especially for deep learning models trained on a narrow staining distribution.

To simulate these realistic distributional shifts, we adopt a biologically grounded perturbation model based on color deconvolution in the Hematoxylin-Eosin-DAB (HED) color space. Inspired by the motivation behind stain-invariant training~\cite{staining}, we introduce this perturbation in the context of test-time evaluation, using controlled shifts in the HED space to model realistic staining variability across clinical settings.

First, each image $\mathbf{x} \in \mathbb{R}^{H \times W \times 3}$ is projected to the Hematoxylin–Eosin–DAB (HED) optical‐density space via a fixed color‐deconvolution matrix $\mathbf{M}_{\mathrm{HED}}$~\cite{ruifrok2001quantification}:
\begin{equation}
\mathbf{s}=\operatorname{vec}\bigl(\mathrm{rgb2hed}(\mathbf{x})\bigr)=\operatorname{vec}\bigl(-\!\log(\mathbf{x}\,\mathbf{M}_{\mathrm{HED}})\bigr)
\in\mathbb{R}^{3HW}
\label{eq:hed_projection}
\end{equation}
where $\operatorname{vec}(\cdot)$ flattens the tensor to a column vector.

Each stain channel $c \in \{\mathrm{H}, \mathrm{E}, \mathrm{D}\}$ is then perturbed independently by a multiplicative and an additive jitter:
\begin{equation}
\mathbf{s}'_{c}=\alpha_{c}\,\mathbf{s}_{c} \,+\, \beta_{c},
\ \ \, 
\alpha_{c} \sim \mathcal{U}(1\!-\!\theta,\,1\!+\!\theta), \;
\beta_{c} \sim \mathcal{U}(-\theta,\,\theta)
\label{eq:hed_jitter}
\end{equation}

where $\theta$ controls corruption severity. We set $\theta\!=\!0.05$ for \textbf{Stain‐Light} and $\theta\!=\!0.2$ for \textbf{Stain‐Heavy}. Finally, the perturbed optical densities $\mathbf{s}'$ are reshaped to $\mathbb{R}^{H\times W\times 3}$, converted back to RGB with $\mathrm{hed2rgb}(\cdot)$, and linearly rescaled to $[0,255]$. A fixed seed is used once to sample $(\boldsymbol{\alpha},\boldsymbol{\beta})$, ensuring deterministic corruptions per image while preventing the model from exploiting inter‐sample consistency.

The intensity scaling controlled by $\alpha_c$ simulates real-world variations such as under- or over-staining, or differences in scanner amplification. Similarly, the additive shifts introduced by $\beta_c$ reflect changes in dye absorption caused by factors like pH, reagent quality, or fixation protocols. Because each of the three stain channels (Hematoxylin, Eosin, and DAB) is perturbed independently, the resulting color changes capture complex and realistic variations in staining that generic color jitter techniques (e.g., HSV or PCA jitter) fail to model accurately. In Section~\ref{sec:base_eval}, we show that even the light setting ($\theta\!=\!0.05$) leads to a substantial drop in the zero-shot performance of a state-of-the-art histopathology VLM, pre-trained on a vast corpus of image–text pairs, underscoring the practical need for robust stain adaptation.



\subsection{Contamination}

Real-world histopathology slides are sometimes affected by physical contaminants introduced during slide preparation, scanning, or storage. As shown in Figure~\ref{fig:motivationm}, artifacts such as dust particles, air bubbles, and tissue folds can obscure diagnostically relevant structures and introduce variability that challenges both pathologists and automated systems~\cite{contamination_image_1, contamination_image_2, contamination_image_3}. These issues are especially critical in digital pathology, where even minor occlusions may compromise tissue interpretation. To simulate such scenarios in a controlled and reproducible way, we design two of the most commonly encountered contamination artifacts: \textbf{Dust} and \textbf{Air Bubble}.

\mypar{Dust.} We model particulate contamination through semi-opaque smudges and fine linear artifacts, commonly seen due to static particles or slide friction. Each dust instance is simulated as a blurred, darkened region using one of two shape types: (1) large rectangular smudges with vertical gradient opacity to mimic streaking or residue, and (2) narrow opaque lines to mimic hairline debris or scratches. A random number of such artifacts (sampled from a uniform range) are placed per image, with Gaussian blur applied to the alpha mask to enhance realism. The final dust mask $M \in [0,1]^{H \times W}$ is applied multiplicatively as an occlusion: 
\begin{equation}
    \mathbf{x}' \,= \, \mathbf{x} \odot (1 - M)
\end{equation}
where $\mathbf{x}$ is the original normalized RGB image.

\mypar{Air Bubble.} We simulate bubble artifacts by overlaying translucent circular regions, combined with local defocus blur and specular highlights. Each bubble is defined by a randomly sampled center and radius, with an alpha-composited RGBA layer simulating light refraction through the bubble’s surface. To enhance realism, we apply circular defocus blur \emph{only} within the bubble region, specified by a binary mask $B\!\in\!\{0,1\}^{H\times W}$. Pixels inside the mask are blurred, while pixels outside remain sharp:
\begin{equation}
    \mathbf{x}' \;=\; (1 - B) \odot \mathbf{x} \;+\; B \odot \operatorname{Blur}_\sigma(\mathbf{x}),
    \label{eq:airbubble}
\end{equation}
where $\operatorname{Blur}_\sigma(\cdot)$ denotes a circular defocus kernel of severity~$\sigma$. Specular highlights are added by drawing brighter inner ellipses and Gaussian-blurring them to mimic reflections, yielding a visually coherent and spatially localized occlusion that resembles micro-bubbles or mounting-medium residue.

\begin{figure*}[h!]
    \centering
    \includegraphics[width=0.85\linewidth]{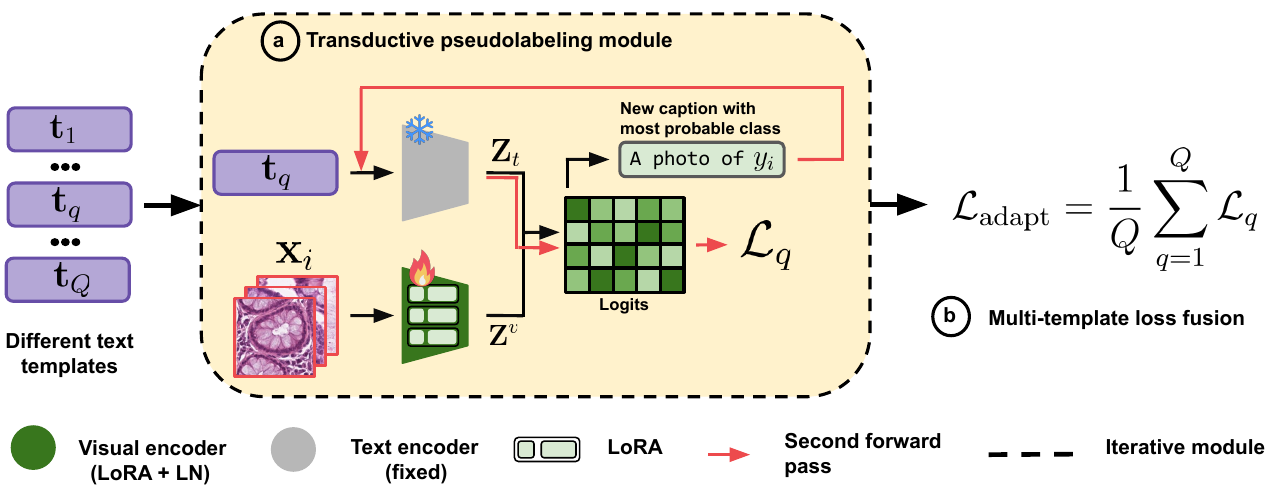}
    \caption{The overall framework of LATTE. All templates are used parallely to compute a loss function. a) We use a transductive pseudolabeling module to align the prediction of each image and its corresponding text caption, and the image-wise and text-wise similarities. b) The losses of the different templates are averaged to build the final adaptation loss used to finetune the model. LoRA and normalization layers are crucial components for adaptation.}
    \label{fig:method}
\end{figure*}

\vspace{-5pt}
\subsection{Blurring}
In addition to staining and contamination artifacts, histopathology images are often affected by acquisition-related blurring (see Figure~\ref{fig:motivationm}), typically caused by imperfect slide preparation, mechanical vibrations, or misaligned focal planes during whole-slide scanning. These degradations obscure fine-grained tissue structures critical for diagnosis and negatively impact both human interpretation and algorithmic performance~\cite{blurring_image,blurring_image_2,blurring_image_3}. To simulate such degradations, we adapt established corruption techniques originally developed for robustness benchmarking in natural images~\cite{cifar10c}, modifying them for use in histopathology. For \textbf{Motion blur}, each image is convolved with a 1D linear kernel of length \(r\) and Gaussian spread \(\sigma\), sampled according to a high severity setting (\(r\!=\!20,\; \sigma\!=\!15\)), and applied at a random direction \(\theta \sim \mathcal{U}(-45^\circ, 45^\circ)\), simulating scanner shake and stage instability. For \textbf{Defocus blur}, a disk-shaped convolutional kernel is applied independently to each color channel, with parameters corresponding to a strong focal distortion (e.g., radius 10, aliasing blur 0.5). These perturbations introduce strong yet plausible degradation to tissue boundaries and cellular features, challenging the spatial precision of visual encoders and providing a rigorous testbed for evaluating test-time adaptation under real-world imaging imperfections.

\subsection{Noise and Illumination}
Finally, we incorporate two additional categories that are less commonly encountered in histopathology but are introduced to induce greater distributional shifts and create more challenging out-of-distribution evaluation scenarios. 

More generally, microscopy images—acquired under varying illumination conditions, sensor types, and photon counts—are often corrupted by signal-dependent shot (Poisson) noise and additive Gaussian noise~\cite{neary2023minimum,kohlberger2019whole,noise_image}, both of which can obscure subtle tissue structures critical for accurate analysis. These degradations have been shown to impair classification performance and image-diagnosis quality~\cite{kohlberger2019whole}. To simulate these effects, we apply \textbf{Gaussian noise} sampled from a standard normal distribution and \textbf{Shot noise} sampled from the Poisson distribution. We inject both at a high-severity setting: Gaussian noise ($\sigma\!=\!0.38$) and shot noise (photon-count factor $c\!=\!3$).

In addition to noise-related degradations, histopathology images may be subject to intensity variations arising from differences in illumination or scanner calibration. Uneven illumination across fields of view, often resembling shading artifacts, can cause intensity distortions that affect downstream analysis~\cite{tak2020simple}. To simulate such effects, we apply two transformations: \textbf{Contrast}, where pixel value differences are scaled to increase or decrease the dynamic range of the image, and \textbf{Brightness}, where an additive constant is applied to all pixels, shifting the overall luminance. Both are applied at a high-severity setting to mimic strong intensity changes—contrast (scaling factor \(= 0.05\)) and brightness (additive shift \(\Delta I = 0.5\)).

Representative examples of the proposed Histopath-C corruptions are illustrated in Figure \ref{fig:benchmark}.

\section{Method}
\label{sec:method}

To introduce our method, we begin by explaining how VLMs are applied to image classification and providing an overview of the TTA setting considered in this work. We identify the essential points leading towards an effective adaptation against histopathology-plausible domain shifts: \emph{a}) a transductive vision-text loss, \emph{b}) low-rank and normalization layers adaptation, and \emph{c}) loss-level text ensembling. 

\mypar{Vision-language classification.} Our method relies on a pre-trained VLM consisting of a visual encoder $f_v$ and a text encoder $f_t$, parameterized by $\theta_v$ and $\theta_t$, respectively. From a batch of $B$ images $\vx_i$ and a set of $C$ classes described using text prompts $\vt_k$, the normalized visual and text features are computed as $\vz_v = f_v (\vx_i)$ and $\vz_t^k = f_t (\vt_k)$. The probability of assigning images to class $c$ is then calculated as
\begin{equation}
    p(y=c\,|\,\vx) \,= \, \frac{\exp{(\vz_v^\top \vz_t^c/\tau)}}{\sum_{k=1}^C \exp{(\vz_v^\top \vz_t^k/\tau)}}
    \label{eq:prediction}
\end{equation}
where $\tau$ is a temperature parameter. The predicted class for an image $\vx_i$ is $\hat{y}_i = \argmax_c p(y\!=\!c\,|\,\vx_i)$. In TTA, the objective is to adapt the model, via $f_v$ or $f_t$, to an unlabeled target dataset $\mathcal{D}_{\mathit{target}} = \{\vx_i\}_{i=1}^N$.

\mypar{Transductive pseudolabeling.} Entropy minimization, the most widely used TTA paradigm, presents severe limitations in VLMs, where using conditional entropy on probabilistic pseudolabels (\eg, softmax logits) tends to cause different degrees of collapse due to overconfidence. This issue is further compounded when multiple text templates are involved, a common scenario in the adaptation of medical VLMs. To address this, we propose integrating both vision and multi-text information to generate more robust pseudolabels, enhancing the adaptation process.

Following~\cite{watt}, we first compute an image-wise similarity matrix $\mS_v = \mZ_v^\top \mZ_v \in \mathbb{R}^{B \times B}$, which quantifies the degree of similarity between each pair of images. Similarly, we use the predicted class label $\hat{y}_i$ of the $i$-th image to generate its corresponding text feature $\hat{\mZ}_{t,i} = f_t(\mathit{template}(\hat{y}_i))$. To capture the similarity between text-based predictions across images, we then compute the text-wise similarity matrix as $\mS_t = \hat{\mZ}_t^\top \hat{\mZ}_t \in \mathbb{R}^{B \times B}$. Finally, the transductive pseudolabels are obtained by combining the image-wise and text-wise similarities as follows:
\begin{equation}
    \hat{\mP}_q \, = \, \mathit{softmax}\Big(\frac{\mS_v + \mS_t}{2}\Big).
    \label{eq:pseudolabels}
\end{equation}
With the new logits $\hat{\mY}^{\text{new}} = \mZ_v^\top \hat{\mZ}_t$, we use the cross-entropy between $\hat{\mY}^{\text{new}}$ and $\hat{\mP}_q$ as our adaptation loss:
\begin{equation}
    \mathcal{L}_q \, = \, \mathcal{H}(\hat{\mY}_q^{\text{new}}, \hat{\mP}_q).
    \label{eq:loss}
\end{equation}
Minimizing this loss, the visual and text encoders are encouraged to output similar embeddings for (\textbf{potentially different}) samples which are related visually or semantically.  


\mypar{Loss-level text ensemble.} To account for the different contribution of templates in solving the classification task, we utilize $Q$ different text templates per task. The loss function in Eq.~\ref{eq:loss} is computed for all text features $\hat{\mZ}_{t,i} = f_t(\mathit{template}_q(\hat{y}_i))$. Finally, the contribution of each template is aggregated on the loss level as a linear combination:
\begin{equation}
    \mathcal{L}_\text{adapt} \, = \, \sum_{q=1}^Q\alpha_q \mathcal{L}_q
    \label{eq:loss_fussion}
\end{equation}

\noindent with $\sum_q \alpha_q\!=\!1$. Besides being a natural solution, a uniformly distributed combination (\ie, $\alpha_q\!=\!1/Q$) demonstrated competitive performance across benchmarks, providing a sufficiently diverse gradient update from the full spectrum of text information. During evaluation, we apply text averaging to derive the final prediction.

\mypar{Adapting beyond normalization layers:} Contrary to the widely established practice of only focusing on normalization layers \cite{tent} for adaptation, we also exploit the recently introduced Low-Rank (LoRA) adaptation \cite{lora}, where the \emph{Queries}, \emph{Keys}, \emph{Values} and MLP layers are fine-tuned through parameter-efficient adapters modeled as the product of two low-rank matrices. A given set of weights $\mW \in \mathbb{R}^{d_{in} \times d_{out}}$ is adapted as $\mW' = \mW + \mB\mA$, with $\mA\in \mathbb{R}^{d_{in} \times r}$ and $\mB\in \mathbb{R}^{r \times d_{out}}$ as matrices of rank $r \ll d_{in}$.

\section{Experiments}
\label{sec:experiments}


\vspace{-6pt}

\mypar{Datasets.} To assess the effectiveness and generalization capabilities of LATTE, we perform extensive evaluations across a suite of diverse and challenging histopathology datasets, encompassing various organs, cancer types, imaging conditions, and annotation granularities, including colorectal cancer (NCT-7K/100K~\cite{nct}), lung/colon adenocarcinoma (LC25000~\cite{lc25000}), skin tumors (SkinCancer~\cite{skincancer}), renal cell carcinoma textures (RenalCell~\cite{renalcell}), and colorectal polyp subtypes (MHIST~\cite{mhist}). Together, these datasets span binary to multi-class classification tasks, fine-grained to coarse-grained labels, and intra-/inter-organ variability. For each, we apply the ten corruptions in Histopath-C, denoted as \emph{{Dataset}-C}. Full dataset details are provided in the supplementary.

\begin{table}[!t]
    \centering
    \resizebox{.5\textwidth}{!}{%
    \begin{tabular}{ll|cccccc}
    \toprule
    \multicolumn{2}{l|}{\begin{tabular}[l]{@{}l@{}}VLM: Quilt\\Dataset\end{tabular}} 
        & Source & TENT & LAME & TPT & CLIPArTT & LATTE \tiny{(Ours)} \\ \midrule
    \multicolumn{2}{l|}{NCT7K}                & 60.86 & 61.65 & 68.91 & 58.00 & 67.01 & \textbf{69.24}\improv{8.4} \\ \midrule
    \parbox[t]{2mm}{\multirow{11}{*}{\rotatebox[origin=c]{90}{NCT7K-C}}} 
    &Stain-Light    & 46.77 & 51.71 & 53.39 & 46.45 & 55.61 & \textbf{65.26}\improv{18.5} \\
    &Stain-Heavy    & 35.72 & 13.18 & 35.85 & 34.29 & 53.99 & \textbf{66.95}\improv{31.2} \\
    &Dust           & 55.53 & 60.86 & 35.85 & 54.47 & 67.80 & \textbf{67.92}\improv{12.4} \\
    &Air Bubble     & 63.89 & 63.19 & 65.01 & 63.19 & 65.02 & \textbf{67.44}\improv{3.6} \\
    &Defocus Blur   & 53.36 & 40.75 & 51.11 & 52.04 & 56.33 & \textbf{64.45}\improv{11.1} \\
    &Motion Blur    & 34.25 & 14.86 & 32.46 & 34.54 & \textbf{49.47} & 46.70\improv{12.5} \\
    &Gaussian Noise & 26.77 & \phz9.59 & 18.25 & 26.91 & 62.92 & \textbf{69.72}\improv{43.0} \\
    &Shot Noise     & 21.59 & 10.11 & 12.18 & 21.41 & 57.63 & \textbf{67.42}\improv{45.8} \\
    &Brightness     & 43.75 & 28.66 & 48.30 & 43.73 & \textbf{57.94} & 57.09\improv{13.3} \\
    &Contrast       & 22.63 & \phz9.35 & 23.92 & 22.52 & 42.85 & \textbf{44.82}\improv{22.2} \\ \cmidrule{2-8}
    & \cellcolor{gray!15}Mean         & \cellcolor{gray!15}40.43 & \cellcolor{gray!15}30.23 & \cellcolor{gray!15}37.63 & \cellcolor{gray!15}39.96 & \cellcolor{gray!15}56.96 & \cellcolor{gray!15}\textbf{61.78}\improv{21.4} \\ \midrule 
    \parbox[t]{2mm}{\multirow{14}{*}{\rotatebox[origin=c]{90}{Other Datasets}}}    
    &{NCT100K}              & 55.98 & 41.42 & 64.06 & 52.83 & 59.86 & \textbf{68.14}\improv{12.2} \\ 
    & \cellcolor{gray!15}NCT100K-C     & \cellcolor{gray!15}39.89 & \cellcolor{gray!15}28.30 & \cellcolor{gray!15}37.20 & \cellcolor{gray!15}38.42 & \cellcolor{gray!15}51.18 & \cellcolor{gray!15}\textbf{56.13}\improv{16.2} \\ \cmidrule{2-8} 
    &{LC25K-Lung}             & 82.87 & 70.34 & 88.00 & 83.03 & -- & \textbf{89.41}\improv{6.5}  \\
    & \cellcolor{gray!15}LC25K-Lung-C         & \cellcolor{gray!15}72.54 & \cellcolor{gray!15}52.65 & \cellcolor{gray!15}74.08 & \cellcolor{gray!15}72.06 & \cellcolor{gray!15}-- & \cellcolor{gray!15}\textbf{78.80}\improv{6.3} \\ \cmidrule{2-8} 
    &{LC25K-Colon}            & 94.41 & 89.28 & 98.70 & 94.50 & -- & \textbf{99.21}\improv{4.8} \\
    & \cellcolor{gray!15}LC25K-Colon-C       & \cellcolor{gray!15}78.13 & \cellcolor{gray!15}55.69 & \cellcolor{gray!15}79.86 & \cellcolor{gray!15}77.58 & \cellcolor{gray!15}-- & \cellcolor{gray!15}\textbf{92.17}\improv{14.0} \\ \cmidrule{2-8} 
    &{LC25K-All}            & 79.28 & 71.39 & \textbf{87.13} & 79.22 & 80.47 & 86.97\improv{7.7} \\ 
    & \cellcolor{gray!15}LC25K-All-C       & \cellcolor{gray!15}57.13 & \cellcolor{gray!15}40.26 & \cellcolor{gray!15}56.61 & \cellcolor{gray!15}56.64 & \cellcolor{gray!15}65.76 & \cellcolor{gray!15}\textbf{71.68}\improv{14.6} \\ \cmidrule{2-8} 
    &{Skin}          & 44.22 & 24.31 & 40.09 & 45.16 & 46.42 & \textbf{50.62}\improv{6.4} \\
    & \cellcolor{gray!15}Skin-C         & \cellcolor{gray!15}22.21 & \cellcolor{gray!15}8.78  & \cellcolor{gray!15}17.30 & \cellcolor{gray!15}22.47 & \cellcolor{gray!15}29.50 & \cellcolor{gray!15}\textbf{33.81}\improv{11.6} \\ \cmidrule{2-8} 
    &{Renal}          & 49.76 & 43.19 & \textbf{50.77} & 50.29 & 43.28 & 46.14\degrad{3.6} \\
    & \cellcolor{gray!15}Renal-C       & \cellcolor{gray!15}30.46 & \cellcolor{gray!15}26.95 & \cellcolor{gray!15}30.40 & \cellcolor{gray!15}30.20 & \cellcolor{gray!15}29.37 & \cellcolor{gray!15}\textbf{38.29}\improv{7.8} \\ \cmidrule{2-8} 
    &{MHIST}                           & 62.95 & 63.15 & 63.10 & 61.51 & -- & \textbf{64.02}\improv{1.1} \\
    & \cellcolor{gray!15}MHIST-C       & \cellcolor{gray!15}57.75 & \cellcolor{gray!15}55.05 & \cellcolor{gray!15}53.29 & \cellcolor{gray!15}57.60 & \cellcolor{gray!15}--    & \cellcolor{gray!15}\textbf{62.01}\improv{4.3} \\ 
    \bottomrule
    \end{tabular}
    }
    \caption{Comparison of test-time adaptation methods with Quilt~\cite{quilt} as the base VLM. Results are reported on multiple datasets under clean and corrupted settings. Gains of our method over the source model are highlighted in green. Note that CLIPArTT is not applicable to datasets with fewer than three classes due to its method constraints. For detailed corruption-specific results and corresponding statistics (mean ± standard deviation over three runs), please refer to the supplementary material.}
    \label{tab:adaptation_comparison}
    \vspace{-6pt}
\end{table}

\begin{table*}[!t]
\centering
\begin{minipage}{0.48\textwidth}
    \centering
    \resizebox{\linewidth}{!}{%
    \begin{tabular}{l|cccccc}
    \toprule
    \begin{tabular}[l]{@{}l@{}}VLM: PathGen\\Dataset\end{tabular} 
        & Source & TENT & LAME & TPT & CLIPArTT & LATTE \tiny{(Ours)} \\ \midrule
    NCT7K & 64.74 & 73.02 & 73.86 & 64.42 & 64.95 & \textbf{79.95}\improv{15.2} \\ 
    \cellcolor{gray!15}NCT7K-C & \cellcolor{gray!15}43.52 & \cellcolor{gray!15}45.07 & \cellcolor{gray!15}44.46 & \cellcolor{gray!15}43.39 & \cellcolor{gray!15}57.51 & \cellcolor{gray!15}\textbf{74.05}\improv{30.5} \\ \midrule
    NCT100K & 66.41 & 67.30 & 68.73 & 65.73 & 69.84 & \textbf{81.73}\improv{15.3} \\ 
    \cellcolor{gray!15}NCT100K-C & \cellcolor{gray!15}45.91 & \cellcolor{gray!15}40.25 & \cellcolor{gray!15}45.53 & \cellcolor{gray!15}45.41 & \cellcolor{gray!15}58.22 & \cellcolor{gray!15}\textbf{71.68}\improv{25.8} \\ \cmidrule{1-7}
    LC25K-Lung & 82.79 & 81.85 & 77.74 & 81.41 & -- & \textbf{96.05}\improv{13.3} \\
    \cellcolor{gray!15}LC25K-Lung-C & \cellcolor{gray!15}67.09 & \cellcolor{gray!15}52.88 & \cellcolor{gray!15}63.94 & \cellcolor{gray!15}66.44 & \cellcolor{gray!15}-- & \cellcolor{gray!15}\textbf{89.37}\improv{22.3} \\ \cmidrule{1-7}
    LC25K-Colon & 95.92 & 97.69 & \textbf{98.98} & 95.82 & -- & 98.27\improv{2.4} \\
    \cellcolor{gray!15}LC25K-Colon-C & \cellcolor{gray!15}72.96 & \cellcolor{gray!15}70.18 & \cellcolor{gray!15}74.07 & \cellcolor{gray!15}72.82 & \cellcolor{gray!15}-- & \cellcolor{gray!15}\textbf{94.39}\improv{21.4} \\ \cmidrule{1-7}
    LC25K-All & 83.25 & 90.47 & 83.74 & 82.38 & 87.40 & \textbf{93.27}\improv{10.0} \\ 
    \cellcolor{gray!15}LC25K-All-C & \cellcolor{gray!15}55.66 & \cellcolor{gray!15}50.00 & \cellcolor{gray!15}56.95 & \cellcolor{gray!15}55.39 & \cellcolor{gray!15}70.52 & \cellcolor{gray!15}\textbf{82.08}\improv{26.4} \\ \cmidrule{1-7}
    Skin & 55.26 & 56.02 & 63.52 & 54.71 & 63.22 & \textbf{68.34}\improv{13.1} \\
    \cellcolor{gray!15}Skin-C & \cellcolor{gray!15}26.22 & \cellcolor{gray!15}18.92 & \cellcolor{gray!15}23.71 & \cellcolor{gray!15}26.18 & \cellcolor{gray!15}38.07 & \cellcolor{gray!15}\textbf{50.65}\improv{24.4} \\ \cmidrule{1-7}
    Renal & 48.61 & 49.66 & 50.28 & 48.67 & 50.42 & \textbf{56.81}\improv{8.2} \\
    \cellcolor{gray!15}Renal-C & \cellcolor{gray!15}23.04 & \cellcolor{gray!15}15.81 & \cellcolor{gray!15}19.59 & \cellcolor{gray!15}22.82 & \cellcolor{gray!15}34.68 & \cellcolor{gray!15}\textbf{44.08}\improv{21.0} \\ \cmidrule{1-7}
    MHIST & 57.11 & 39.00 & 46.21 & 57.22 & -- & \textbf{60.49}\improv{3.4} \\
    \cellcolor{gray!15}MHIST-C & \cellcolor{gray!15}55.29 & \cellcolor{gray!15}46.08 & \cellcolor{gray!15}49.54 & \cellcolor{gray!15}54.94 & \cellcolor{gray!15}-- & \cellcolor{gray!15}\textbf{56.22}\improv{0.9} \\ 
    \bottomrule
    \end{tabular}}
    \caption{Comparison of test-time adaptation methods with PathGen~\cite{pathgen} as the base VLM. Results are reported on multiple datasets under clean and corrupted settings. Gains of our method over the source model are highlighted in green.}
    \label{tab:adaptation_comparison_pathgen}
\end{minipage}%
\hfill
\begin{minipage}{0.48\textwidth}
    \centering
    \resizebox{\linewidth}{!}{%
    \begin{tabular}{l|cccccc}
    \toprule
    \begin{tabular}[l]{@{}l@{}}VLM: CONCH\\Dataset\end{tabular} 
        & Source & TENT & LAME & TPT & CLIPArTT & LATTE \tiny{(Ours)} \\ \midrule
    NCT7K & 67.55 & 68.66 & 73.59 & 67.41 & 58.28 & \textbf{82.12}\improv{14.6} \\
    \cellcolor{gray!15}NCT7K-C & \cellcolor{gray!15}37.35 & \cellcolor{gray!15}38.19 & \cellcolor{gray!15}38.66 & \cellcolor{gray!15}37.29 & \cellcolor{gray!15}39.09 & \cellcolor{gray!15}\textbf{67.91}\improv{30.6} \\ \midrule
    NCT100K & 63.79 & 65.00 & 71.82 & 63.73 & 59.66 & \textbf{82.76}\improv{19.0} \\ 
    \cellcolor{gray!15}NCT100K-C & \cellcolor{gray!15}32.98 & \cellcolor{gray!15}32.32 & \cellcolor{gray!15}34.24 & \cellcolor{gray!15}32.94 & \cellcolor{gray!15}37.01 & \cellcolor{gray!15}\textbf{63.53}\improv{30.6} \\ \cmidrule{1-7}
    LC25K-Lung & 87.95 & 90.22 & 92.67 & 87.45 & -- & \textbf{97.02}\improv{9.1} \\
    \cellcolor{gray!15}LC25K-Lung-C & \cellcolor{gray!15}56.13 & \cellcolor{gray!15}52.00 & \cellcolor{gray!15}55.95 & \cellcolor{gray!15}56.17 & \cellcolor{gray!15}-- & \cellcolor{gray!15}\textbf{83.72}\improv{27.6} \\ \cmidrule{1-7}
    LC25K-Colon & 95.31 & 96.32 & 97.46 & 95.15 & -- & \textbf{99.18}\improv{3.9} \\
    \cellcolor{gray!15}LC25K-Colon-C & \cellcolor{gray!15}71.19 & \cellcolor{gray!15}69.93 & \cellcolor{gray!15}71.57 & \cellcolor{gray!15}71.10 & \cellcolor{gray!15}-- & \cellcolor{gray!15}\textbf{97.42}\improv{26.2} \\ \cmidrule{1-7}
    LC25K-All & 86.58 & 87.98 & 90.55 & 86.17 & 89.49 & \textbf{95.81}\improv{9.2} \\ 
    \cellcolor{gray!15}LC25K-All-C & \cellcolor{gray!15}51.17 & \cellcolor{gray!15}49.26 & \cellcolor{gray!15}51.98 & \cellcolor{gray!15}51.17 & \cellcolor{gray!15}48.06 & \cellcolor{gray!15}\textbf{80.38}\improv{29.2} \\ \cmidrule{1-7}
    Skin & 34.56 & 33.34 & 35.98 & 34.11 & 31.39 & \textbf{64.32}\improv{29.8} \\
    \cellcolor{gray!15}Skin-C & \cellcolor{gray!15}21.52 & \cellcolor{gray!15}19.64 & \cellcolor{gray!15}21.51 & \cellcolor{gray!15}21.44 & \cellcolor{gray!15}21.85 & \cellcolor{gray!15}\textbf{47.61}\improv{26.1} \\ \cmidrule{1-7}
    Renal & 46.16 & 48.74 & 51.26 & 45.65 & 52.87 & \textbf{57.22}\improv{11.1} \\
    \cellcolor{gray!15}Renal-C & \cellcolor{gray!15}20.62 & \cellcolor{gray!15}18.16 & \cellcolor{gray!15}20.21 & \cellcolor{gray!15}20.58 & \cellcolor{gray!15}22.64 & \cellcolor{gray!15}\textbf{48.99}\improv{28.4} \\ \cmidrule{1-7}
    MHIST & 59.98 & 62.59 & 60.39 & 59.72 & -- & \textbf{63.00}\improv{3.0} \\
    \cellcolor{gray!15}MHIST-C & \cellcolor{gray!15}57.89 & \cellcolor{gray!15}57.45 & \cellcolor{gray!15}\textbf{58.78} & \cellcolor{gray!15}58.06 & \cellcolor{gray!15}-- & \cellcolor{gray!15}56.59\degrad{1.3} \\ 
    \bottomrule
    \end{tabular}}
    \caption{Comparison of test-time adaptation methods with CONCH~\cite{conch} as the base VLM. Results are reported on multiple datasets under clean and corrupted settings. Gains of our method over the source model are highlighted in green.}
    \label{tab:adaptation_comparison_conch}
\end{minipage}
\end{table*}

\mypar{Baselines.} We follow the well established practices in TTA, and incorporate popular methods into our baseline evaluation. We use TENT~\cite{tent}, the most general entropy minimization method, TPT~\cite{tpt} for prompt tuning, LAME~\cite{lame} as a parameter-free method, and CLIPArTT~\cite{clipartt} as a VLM-oriented method. Adaptation is performed for 10 iterations, with a learning rate of $10^{-3}$ on the Adam optimizer, applied on batches of 128 images. We report results across three pathology VLMs—Quilt~\cite{quilt}, PathGen~\cite{pathgen}, and CONCH~\cite{conch}. Quilt uses ViT-B/32 at $224{\times}224$; PathGen uses ViT-B/16 at $224{\times}224$; CONCH uses ViT-B/16 at $448{\times}448$. Unless otherwise stated (e.g., in ablations), Quilt serves as the default VLM.

\subsection{Baseline evaluation}
\label{sec:base_eval}

\mypar{Performance on Histopathology Images.} Different pathology VLMs rely on distinct backbones, tokenization schemes, and training strategies. To comprehensively evaluate Histopath-C, we report results separately for Quilt, PathGen, and CONCH across datasets and adaptation methods in Tables~\ref{tab:adaptation_comparison}, \ref{tab:adaptation_comparison_pathgen}, and \ref{tab:adaptation_comparison_conch}, respectively. 

\noindent Recent state-of-the-art TTA methods based on entropy minimization, such as TENT and TPT, can lead to performance degradation on certain datasets, notably on NCT-100K, as shown in the Table~\ref{tab:adaptation_comparison}. This suggests that confidence-based adaptation strategies are not consistently reliable in the context of histopathology. In particular, TENT, which often serves as a strong baseline for TTA in natural image benchmarks, frequently degrades performance in our setting. These findings indicate that entropy minimization applied to normalization layers may be ill-suited for histopathology images, potentially due to the high intra-class variability and subtle structural differences characteristic of this domain. In contrast, pseudo-labeling approaches, such as CLIPArTT and our proposed method, LATTE, demonstrate significantly greater robustness. Notably, LAME also achieves strong performance without requiring backpropagation. However, LATTE consistently outperforms all methods across many evaluated datasets, improving performance by approximately 9\% over the baseline for NCT7K, for instance.

\noindent Beyond Quilt, we validate our findings on PathGen (Table~\ref{tab:adaptation_comparison_pathgen}) and CONCH (Table~\ref{tab:adaptation_comparison_conch}), two recent pathology-specific VLMs with distinct backbones and training pipelines. Both models show similar trends: entropy-based methods (TENT, TPT) often degrade performance, while pseudo-labeling and parameter-free approaches are more robust. Notably, although CONCH achieves strong accuracy on clean datasets, it degrades more severely under corruptions—an interesting robustness gap. Across datasets and corruption types, LATTE consistently achieves the largest improvements, in many cases exceeding 20–30\% over the source baseline.

\mypar{Performance on Real-World Corruptions.}
We apply the proposed corruptions to all datasets, leading to a performance drop for most methods compared to the baseline. Interestingly, in the case of NCT7K, applying the \emph{Air Bubble} corruption unexpectedly improves performance over the baseline in Table~\ref{tab:adaptation_comparison}, suggesting that Quilt may have encountered similar artifacts during pretraining on colorectal cancer images.
Our proposed method, LATTE, consistently improves performance across all corruptions and datasets compared to the baseline. On average, LATTE also outperforms other TTA methods. However, it can be less effective on specific corruptions, such as \emph{Motion Blur} and \emph{Brightness} on NCT7K, where CLIPArTT achieves slightly better results. Nevertheless, LATTE remains highly robust across different corruptions, achieving substantial improvements of approximately 10\% to 20\% over CLIPArTT and the baseline for \emph{Stain-Light} for example. Moreover, it achieves an average performance improvement of at least 4\% over the second-best method across the corrupted datasets.

\begin{figure*}[!t]
    \centering   
    \begin{minipage}{0.35\textwidth}
        \centering
        \includegraphics[width=0.80\linewidth]{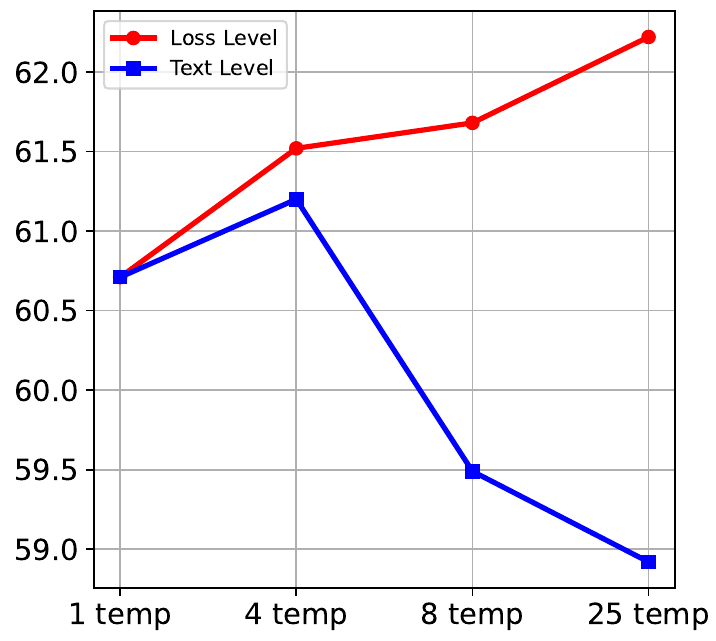}
        \caption{Comparison of Text and Loss averaging over several templates.}
        \label{fig:lossvstext}
    \end{minipage}%
    \hfill
    \begin{minipage}{0.61\textwidth}
        \centering
            \begin{tabular}{ll|ccc}
    \toprule
    \multicolumn{2}{l|}{\begin{tabular}{@{}l@{}}VLM: Quilt \\ Dataset\end{tabular}} & LNs & LoRA & LoRA\,+\,LNs \\ \midrule
    \multicolumn{2}{l|}{NCT7K}  & 69.42\ppm0.02  & 69.63\ppm0.40  & \textBF{69.93\ppm0.18}  \\ \midrule
    \parbox[t]{2mm}{\multirow{6}{*}{\rotatebox[origin=c]{90}{NCT7K-C}}} 
    &Defocus Blur   & 60.00\ppm0.31  & 63.01\ppm0.36  & \textBF{63.38\ppm0.03}  \\
            &Contrast       & \textBF{43.74\ppm0.16}  & 41.52\ppm0.28  & 43.40\ppm0.48  \\
            &Stain-Light    & 58.23\ppm0.24  & 60.82\ppm0.09  & \textBF{60.90\ppm0.13}  \\
            &Stain-Heavy    & 62.66\ppm0.09  & 65.23\ppm0.02  & \textBF{65.96\ppm0.39}  \\
            &Dust           & 69.72\ppm0.28  & \textBF{70.05\ppm0.21}  & 69.91\ppm0.15  \\ \cmidrule{2-5} 
            & \cellcolor{gray!15}Mean  & \cellcolor{gray!15}58.87          & \cellcolor{gray!15}60.13          & \cellcolor{gray!15}\textBF{60.71}  \\ \bottomrule
            \end{tabular}
        \captionsetup{type=table}
        \caption{Comparison on the updated parameters.}
        \label{tab:updatedparameters}
    \end{minipage}%
\end{figure*}

\begin{figure*}[h!]
    \centering   
    \begin{minipage}{0.6\textwidth}
        \centering
            \begin{tabular}{ll}
                \toprule
                ~ & Template\\ \midrule
                $T^1$: & ``\texttt{a histopathology slide showing \{class $k$\}}'' \\
                $T^2$: & ``\texttt{histopathology image of \{class $k$\}}'' \\
                $T^3$: & ``\texttt{pathology tissue showing \{class $k$\}}'' \\
                $T^4$: & ``\texttt{presence of \{class $k$\} tissue on image}'' \\
                $T^5$: & ``\texttt{a photomicrograph showing \{class $k$\}}'' \\
                $T^6$: & ``\texttt{a photomicrograph of \{class $k$\}}'' \\
                $T^7$: & ``\texttt{an image of \{class $k$\}}'' \\
                $T^8$: & ``\texttt{an image showing \{class $k$\}}'' \\
                \bottomrule\\[-8pt]
            \end{tabular}
        \captionsetup{type=table}
        \caption{Some examples of the used text templates.The complete list is provided in the supplementary material.}
        \label{tab:Templates}
    \end{minipage}%
    \hfill
    \begin{minipage}{0.35\textwidth}
        \centering
        \includegraphics[width=0.80\linewidth]{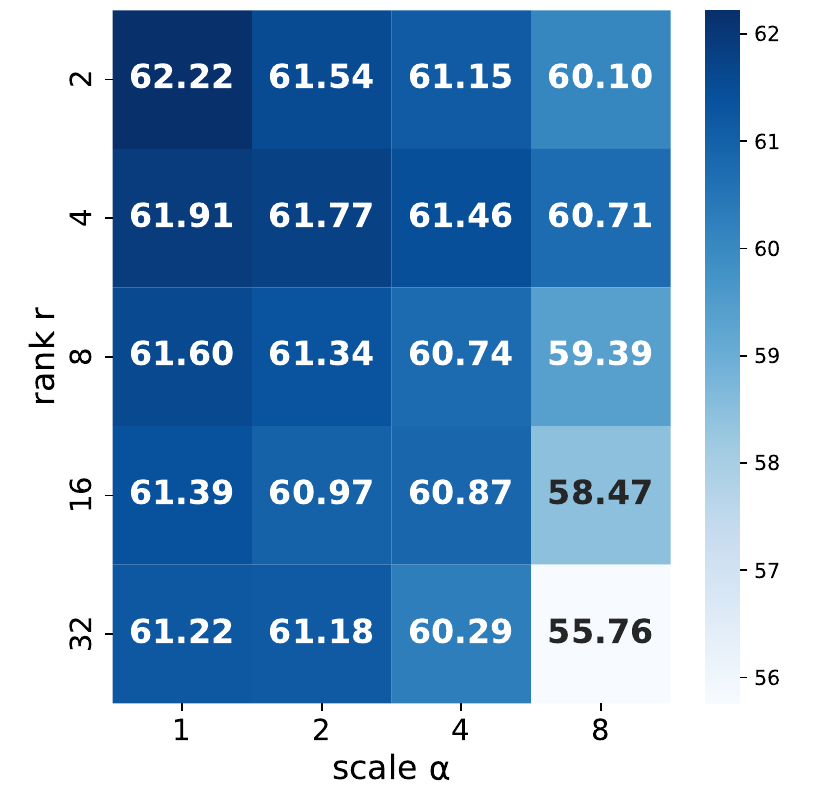}     
        \caption{Comparison of rank $r$ and scale $\alpha$ to leverage LoRA.}
        \label{fig:loraparameters}
    \end{minipage}%
\end{figure*}


\subsection{Ablation Studies}

\mypar{On the updated parameters.} We investigate the optimal strategy for updating our model in the context of test-time adaptation. Specifically, we compare three approaches: updating only the normalization layers, as commonly adopted in prior TTA studies; updating only the attention mechanism using LoRA; and updating both components simultaneously. As shown in Table~\ref{tab:updatedparameters}, the effectiveness of each strategy varies depending on the type of corruption. For instance, updating only the normalization layers proves beneficial for \emph{Contrast} distortions, whereas leveraging LoRA is preferable for \emph{Dust} perturbations. However, in a broader sense, jointly updating both LoRA and normalization layers tends to yield more accurate results across a wider range of corruptions.


\mypar{Text averaging vs Loss averaging.}
As discussed in Section~\ref{sec:intro}, leveraging multiple templates is crucial due to the nature of the textual data used to train medical VLMs. Figure~\ref{fig:lossvstext} demonstrates that employing multiple templates yields better performance compared to using a single one. Specifically, using four templates improves accuracy, whether through text averaging or loss averaging. However, when increasing the number of templates further, performance degrades during adaptation with text averaging, whereas loss averaging continues to improve results. Consequently, we adopt loss averaging as our preferred strategy, using a set of 25 templates (examples of which are provided in Table~\ref{tab:Templates}).





\mypar{LoRA parameters.} We investigate the optimal hyperparameters for our LoRA-based adaptation strategy by evaluating different scaling factors $\alpha$ and ranks $r$, as presented in Figure~\ref{fig:loraparameters}. The results indicate that smaller values of $\alpha$ and $r$ lead to improved performance. Specifically, setting $\alpha\!=\!1$ and $r\!=\!2$ optimizes the LoRA-based strategy, yielding the best adaptation performance.

\subsection{Discussion and limitations}

This paper explores TTA not only as a tool to safely deploy medical VLMs in histopathology, but also as a necessary paradigm for the application of deep models. In that spirit, our benchmark, Histopath-C, is one of the first of its type to emulate real-world shifts, and to severely challenge previous adaptation methods. We recognize, however, that the obtained performance represents the first steps towards a clinically-ready application. 

We evaluate LATTE as a simple, yet effective ensembling approach for TTA in the medical field. We show that combining multiple text semantics using different loss signals, provide a more positive feedback to the model (coupled with LoRA). For instance, our technique outperforms the major exponents of TTA per category; entropy-based (TENT), parameter-free (LAME), prompt tuning (TPT), and transductive adaptation (CLIPArTT). 
\section{Conclusion}
\label{sec:conclusions}
In this work, we introduce Histopath-C, a new benchmark featuring 10 corruption types applicable to any histopathology dataset to simulate real-world distribution shifts. We also present the first test-time adaptation framework for vision-language models in this domain, evaluating recent TTA methods under these challenging conditions. To enhance adaptation, we propose LATTE a novel approach that integrates loss averaging across multiple templates, transductive pseudo-labeling, and low-rank adaptation. We further introduce a diverse set of 25 templates to support generalization to medical reports. Through an extensive ablation study, we provide deeper insights into our method’s design choices and their impact. Comparative evaluations across multiple histopathology datasets demonstrate the superiority of our approach across all scenarios. To further advance realistic adaptation settings, future work could explore batch-based adaptation with mixed corruption types.

{
    \small
    \bibliographystyle{ieeenat_fullname}
    \bibliography{main}
}

\end{document}